\documentclass[letterpaper, 10 pt, conference]{ieeeconf}

\IEEEoverridecommandlockouts                              

\usepackage{graphicx}
\usepackage{amsmath}
\usepackage{xcolor}
\usepackage{floatrow}
\usepackage{epstopdf}
\usepackage[colorlinks = true,
            linkcolor = black,
            urlcolor  = black,
            citecolor = black,
            anchorcolor = black]{hyperref}

\usepackage{algorithm,algpseudocode}
\usepackage[font=small, tableposition=top]{caption}
\usepackage{accents}
\usepackage{amsfonts}
\usepackage{wrapfig}
\usepackage{booktabs}
\usepackage{balance}
\usepackage{scalerel}
\usepackage{subfigure}
\usepackage{multirow}
\usepackage{url}
\usepackage{svg}
\usepackage{hhline}
\usepackage{hyperref}

\long\def\ignore#1{}
\begin{document}
\title{\LARGE \bf
Manipulation-Oriented Object Perception in Clutter \\ through Affordance Coordinate Frames
}

\author{Xiaotong Chen,
        Kaizhi Zheng,
        Zhen Zeng,
        Cameron Kisailus,
        Shreshtha Basu,  \\
        James Cooney,
        Jana Pavlasek,
        Odest Chadwicke Jenkins%
%
\thanks{
X. Chen, K. Zheng, Z. Zeng, C. Kisailus, S. Basu, J. Cooney, J. Pavlasek, and O. C. Jenkins are with the Department of Electrical Engineering and Computer Science, and Robotics Institute at the University of Michigan, Ann
Arbor, MI, USA {\tt\footnotesize \{cxt, zhengkz, zengzhen, kisailus, shreyb, cooneyj, pavlasek, ocj\}@umich.edu}}%
}


\maketitle

\begin{abstract}

In order to enable robust operation in unstructured environments, robots should be able to generalize manipulation actions to novel object instances. 
For example, to pour and serve a drink, a robot should be able to recognize novel containers which afford the task. Most importantly, robots should be able to manipulate these novel containers to fulfill the task.
To achieve this, we aim to provide robust and generalized perception of object affordances and their associated manipulation poses for reliable manipulation. In this work, we combine the notions of affordance and category-level pose, and introduce the Affordance Coordinate Frame (ACF). With ACF, we represent each object class in terms of individual affordance parts and the compatibility between them, where each part is associated with a part category-level pose for robot manipulation. In our experiments, we demonstrate that ACF outperforms state-of-the-art methods for object detection, as well as category-level pose estimation for object parts. We further demonstrate the applicability of ACF to robot manipulation tasks through experiments in both simulation and real world environment.

\end{abstract}


\section{Introduction}
\begin{figure}
    \centering
    \includegraphics[width=.9\columnwidth]{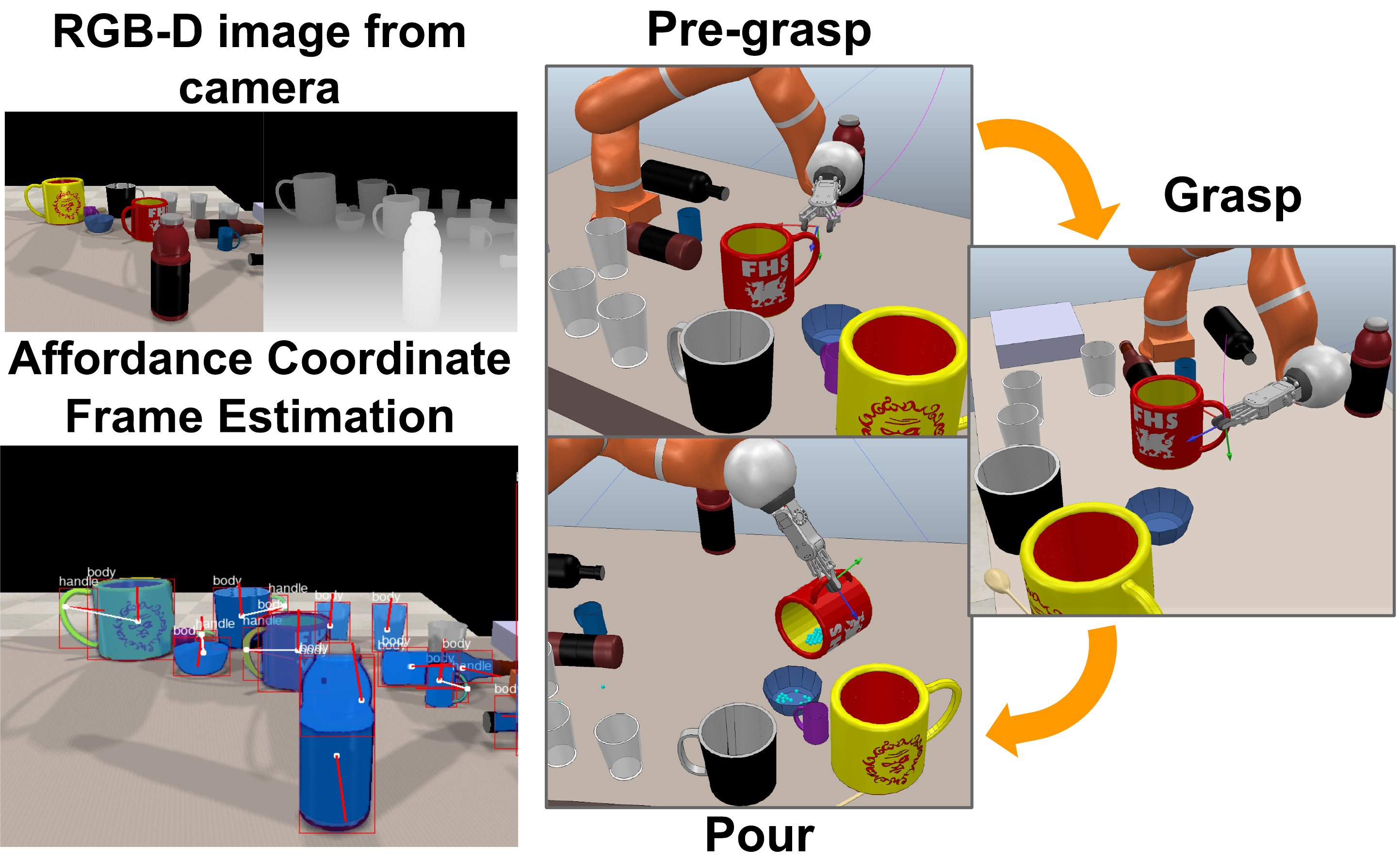}
    \caption{Affordance Coordinate Frames (ACF) provide keypoint and axis constraints for each manipulation action afforded by an object. In the above example, we estimate a ``grasp'' ACF on the handle of a mug, as well as a ``pour'' ACF on the body of a mug (bottom left). The robot uses the estimated ACFs to execute a task involving pouring liquid from the mug (right). The ACF is generalizable across object instances.}
    \label{fig:teaser}
\end{figure}

\begin{figure}
    \centering
    \includegraphics[width=\columnwidth]{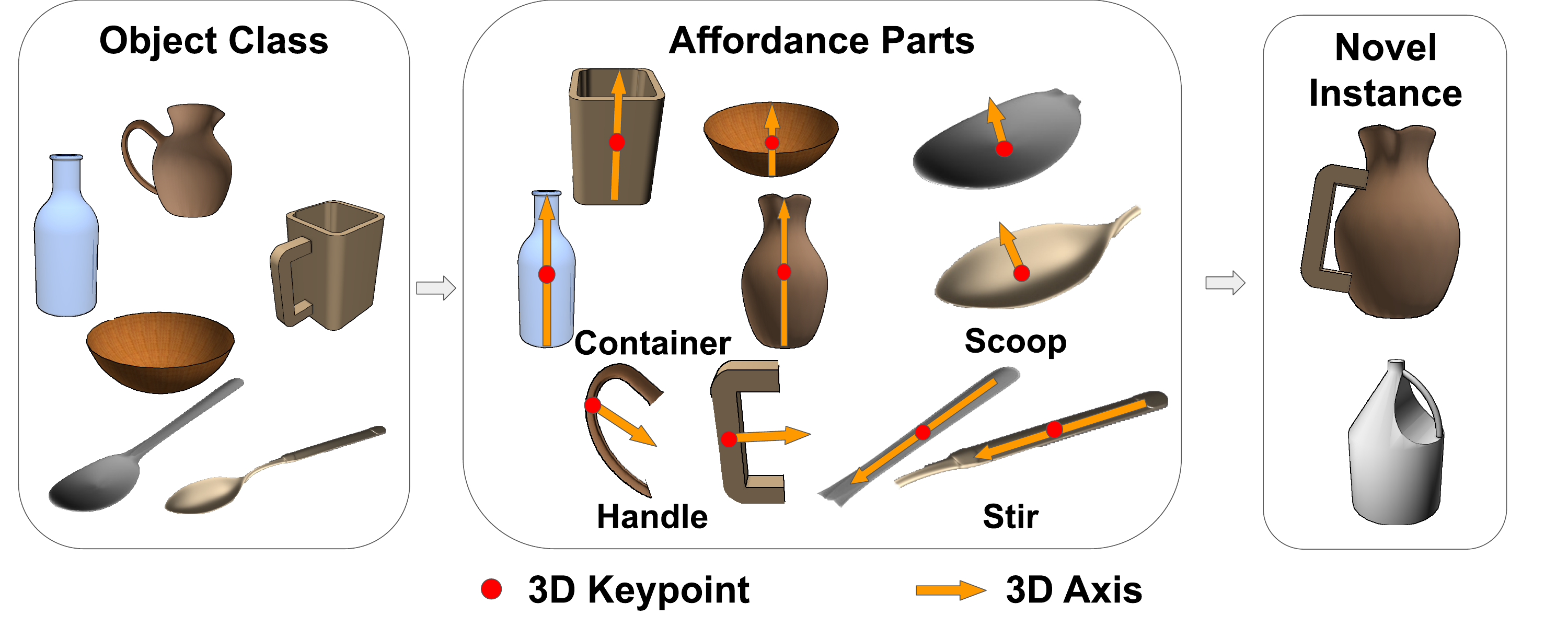}
    \caption{Objects can be divided into functional parts which represent their affordances. The Affordance Coordinate Frame (ACF) is defined as a 3D keypoint and a directed axis, which are consistent across affordance classes. The ACF can be used by a robot to plan manipulation actions. The object parts can be re-assembled to generate new instances with novel shapes. 
    }
    \label{fig:acf_explain}
\end{figure}

In order for robots to assist people in unstructured environments such as homes and hospitals, they must be capable of performing diverse manipulation actions.
Simple pick and place actions are not sufficient for robots to exploit the affordances~\cite{gibson1966senses} of objects when executing common tasks, such as grasping a container to pour as shown in Figure~\ref{fig:teaser}.
Object affordances~\cite{varadarajan2012afrob} provide an agent-centric way of describing available actions that an object offers given the capability of the agent. To interact with common household objects, a robot must make use of the actions afforded by the component parts of the object. 
In this work, we exploit the notion of affordance and represent an object class as a composition of functional parts, which we call \textit{affordance parts}. As shown in Figure~\ref{fig:acf_explain}, the concept of a mug can be generalized to the composition of two affordance parts, namely the container and the handle parts. 


Recently, the notion of affordance for manipulation has garnered increasing interest in the robotics community. Previous work on object affordances for robotic manipulation focuses on recognizing the affordance labels of novel objects~\cite{do2018affordancenet}, which does not directly translate to specific manipulation poses for use by a planner towards task completion. Recent advances have enabled pose estimation of novel objects at the category level~\cite{wang2019normalized, chen2020learning}. However, robots often need fine-grained pose information about specific object parts (e.g. a container handle) for manipulation actions. For such tasks, the category-level object pose is not sufficient to perform manipulation, due to large intra-category variance. Generalizable representations for object localization using keypoints have been proposed~\cite{manuelli2019kpam, qin2019keto}, but they rely heavily on user-specified constraints in order to define manipulation actions over the keypoints.


In this paper, we aim to develop a generalized object representation, the Affordance Coordinate Frame (ACF), that seamlessly links object part affordances and robot manipulation poses. Our key insight lies in that each object class is composed of a collection of functional parts, where each part is strongly associated with its affordances (see Figure~\ref{fig:acf_explain}). For robots to exploit the affordances of object parts, we learn a category-level pose for each object part, similarly to the notion of category-level pose for objects~\cite{wang2019normalized}. In particular, the category-level pose for each object part serves as a coordinate frame (hence the name Affordance Coordinate Frame) in which to pre-define manipulation poses, such as grasp and pour actions.
Building on the insights from Zeng et al.~\cite{zeng2019unsupervised}, our work proposes a deep learning based pipeline for estimating the ACF of object parts from RGB-D observations.
A robot can attach pre-defined manipulation poses to the estimated ACF when executing tasks. By defining the ACF with respect to object parts, our method allows for better generalization and robustness under occlusion, because the robot can see and act on observable parts directly. For example, the handle of a mug can be grasped when its container part is not visible. In contrast to works aimed at pose estimation for robotic manipulation~\cite{tremblay2018deep, pavlasek2020parts}, we do not assume known 3D object geometry models or 6D object poses. 
We show the accuracy and generalization of our method in detecting novel object instances, as well as estimating ACF for novel object parts in our experiments. We demonstrate that the proposed method outperforms the state-of-the-art methods for object detection, as well as category-level pose estimation for object parts. We then demonstrate manipulation tasks through ACF estimation in cluttered environments in both simulation and on a Fetch manipulation platform.


\section{Related Work}
In the following subsections, we discuss related works in three aspects: the \textit{affordance} concept initiates the proposition of our work that connects visual perception and robotic manipulation through our proposed representation; the \textit{object pose estimation} basically shows a pipeline of manipulation based on poses of entire objects, with which we compare the estimation accuracy in experiments; the \textit{task-oriented grasping} includes related application to robot grasping, with which we both interact with objects considering part semantics while our pipeline also fits other actions than grasp.

\subsection{Affordances for Robotic Manipulation}

The concept of affordance has been explored in many aspects of robot manipulation. For visual perception, Nguyen et al.~\cite{do2018affordancenet} proposed AffordanceNet to achieve state-of-the-art accuracy on affordance detection and segmentation on affordance datasets~\cite{nguyen2017object,myers2015affordance}. 
The affordance detection together with segmentation can reduce search space of grasp pose for manipulation as explored in~\cite{detry2017task}, and grasp center point and axis are fitted from the shape of masks in~\cite{chu2019learning}. Our main contribution is to propose a model that jointly learns part-based 3D keypoints and axes as well as association between parts.
On the other hand, various affordance representations have been proposed for task execution. Affordance templates~\cite{hart2015affordance} were developed for highly constrained manipulation tasks but require user inputs to manually register it with robot observations.
Affordance wayfield~\cite{mcmahon2018affordance} extended the similar idea to a gradient field.
In~\cite{kaiser2016towards, kaiser2018autonomous}, affordances are modeled as an existence possibility using certainty or belief functions, based on the relation of environmental shape primitives and end-effector poses. The formulation supports hierarchical relations from basic grasps to bi-manual operations.
Our work differs from these works as we define affordances from object part level to be visually perceivable and generalizable to objects with similar functionality.

Recent works that connect visual perception of object affordances with robot manipulations are most relevant to our work. kPAM~\cite{manuelli2019kpam} uses 3D semantic keypoints as the object representation for category-level object manipulations. The core advantage of ACF over kPAM is: when multiple instances are presented in the cluttered environment, ACF has the notion of compatibility to determine which ACFs compose to the same object, so it would be robust to cluttered scenes.
Specific to tool manipulation, Qin et al.~\cite{qin2019keto} also proposed a keypoint-based representation, which specifies the grasp point and function point (hammer head for example) on the tool and effect point on the target object. The robot motion was solved through optimization over the keypoint-based constraints. Compared to 3D keypoints registered on whole objects, our ACF representation decomposes object into object parts based on their functionality, and include explicit orientation for manipulation. Similarly, kPAM 2.0~\cite{gao2021kpam} and AffKp~\cite{xu2021affordance} also introduces the orientation from/with keypoints to define manipulation constraints.


\subsection{Object Pose Estimation for Manipulation}
Object pose estimation offers a formal way to perform manipulation with known object geometry models. Instance-level pose estimators have been extensively studied in the vision community~\cite{tremblay2018deep,he2020pvn3d}. 
In the robotics community, works have focused on the problem of object localization in clutter for the purpose of robotic pick and place tasks~\cite{zeng2018srp, sui2015axiomatic, chenGrip}. 
The success of parts-based methods have been demonstrated in highly cluttered scenes~\cite{Desingheaaw4523, pavlasek2020parts}. 
However, they rely on known mesh models and do not extensively validate the representations for manipulation.

Recently, several works aim to extend that to category-level pose estimation, which relaxes the assumption of known object geometry models. Wang et al.~\cite{wang2019normalized} proposed Normalized Object Coordinate Space (NOCS) as a dense reconstruction of object category in canonical space, under the assumption that the intra-category shape variance can be well approximated by a representative 3D shape. 
Chen et al.~\cite{chen2020learning} gave another category-level representation modeled by a latent space vector generated from a variational autoencoder, which does not assume dense point correspondence. 
A recent work~\cite{li2020category} extended~\cite{wang2019normalized} to articulated objects, which solves the joint axes and parameters
after estimating poses of individual parts. 
Compared to these representations of whole objects, the ACF focuses on functional parts of objects rather than entire object-level poses.


\subsection{Task-Oriented Grasping}

Task-oriented grasping incorporates semantics into classic grasping to provide richer affordance-based manipulation.
Our part-based affordance representation shares the same idea as in~\cite{detry2017task, kokic2017affordance, liu2019cage} that use task semantics to prioritize different regions of objects for task-oriented grasping detection. Similarly in~\cite{song2020robust}, task-based external wrench is considered through grasp analysis to extend DexNet~\cite{mahler2019learning} to a task-oriented grasping service. Compared to~\cite{antanas2019semantic}, our representation also divides the whole objects into functional parts but also adds keypoints and axes for ease of manipulation.
In general, we share the common part-based representation for robot manipulation, while ACFs can afford guidance on the motion trajectory after grasping by modeling the whole objects into connected parts.

\section{Affordance Coordinate Frames}
\begin{figure*}
    \centering
    \includegraphics[width=0.8\columnwidth]{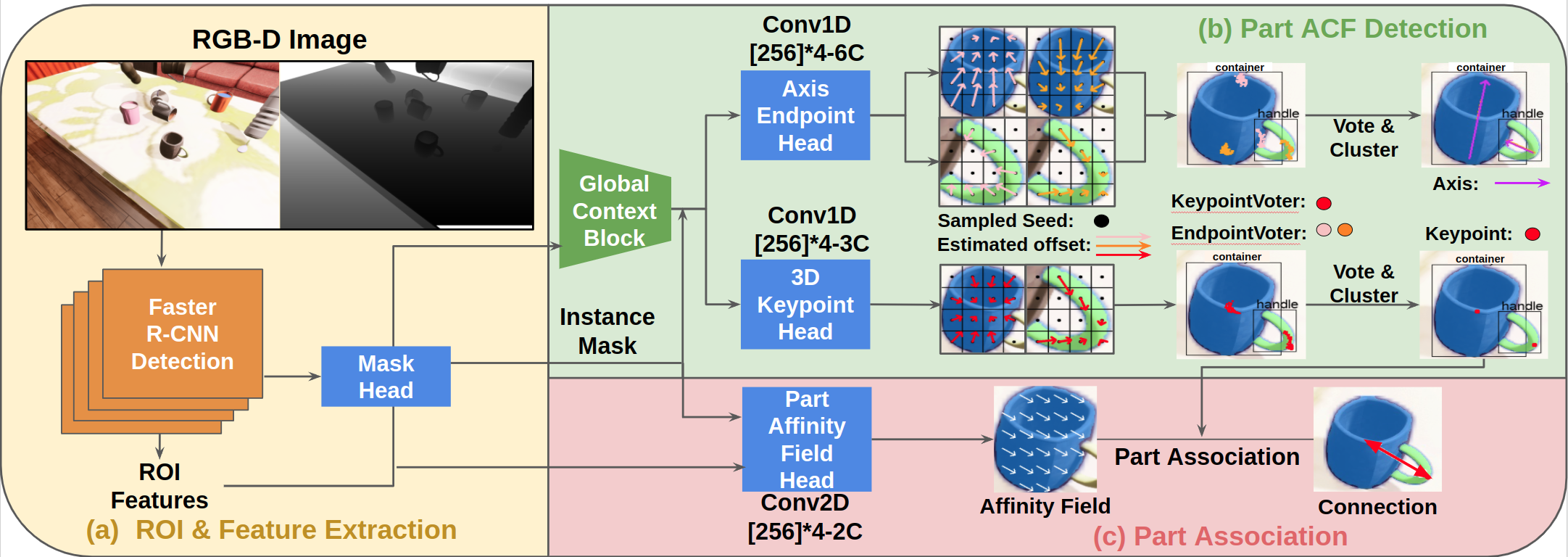}
    \caption{The perception pipeline for estimating the affordance coordinate frames of a mug with two affordance parts: a container part and a handle part. (a) The Faster R-CNN module detects bounding boxes from features extracted through feature pyramid network out of a 4-channel fused RGB-depth image. (b) The corresponding region of interest (ROI) features are further fused with global features using self-attention module in~\cite{cao2019GCNet}. 3D keypoint and directed axis endpoint offsets are learned in two parallel network heads ($C$ is the number of part classes). The voters are formed by adding the learned offsets to uniformly sampled seeds within a bounding box and estimated keypoints and axes are computed by clustering the voter points. (c) Another network head learns 2D part affinity fields, which connect parts of an object.}
    \label{fig:framework}
\end{figure*}

\subsection{Representation} 

Given a predefined task, our goal is to detect and localize objects in a scene in terms of their parts such that manipulation actions can be executed towards task completion. The problem is divided into two subtasks: the first is to perceive affordances from RGB-D observations from a robot, and the second is to generate manipulation actions from the perceived affordances. 
We define affordances as the functional interaction between an object and an agent. 
For common household objects, such affordances are associated with object parts, rather than the whole object, and we therefore associate ACFs with object parts.
An ACF is formally defined as a 3D keypoint with a directed axis with its origin at the keypoint location.
The ACF allows the affordance to be perceived from visual data while also acting as a concrete and generalization parametrization of a robot action. 
Figure~\ref{fig:acf_explain} illustrates the geometric relation of the defined axes. The advantages of the ACF are two-fold: 
\begin{enumerate}
    \item Better generalization to novel objects with similar functionality, since parts in one category have smaller visual difference compared to full objects.
    \item They provide a sparse representation while providing enough constraints for robotic manipulation by exploiting the symmetric nature of object parts.
\end{enumerate}

\subsection{Estimation Pipeline}


We estimate instance-level object part ACFs using a network similar to Mask R-CNN~\cite{he2017mask}.
The overall framework is illustrated in Figure~\ref{fig:framework}. To incorporate depth information, we add an additional channel to the first backbone layer to accept 4-channel RGB-D input. We add three network head architectures for 3D keypoint, axis and part affinity field estimation. Similar to the mask head, estimates are made in every proposed region of interest (ROI) with a detected bounding box and corresponding ROI feature.
We concatenate each ROI feature with a global feature computed using Global Context Block (a self-attention module) from~\cite{cao2019GCNet} to fuse spatial and channel context. 
In this pipeline, all three heads use convolutional layers and we employ deep Hough voting~\cite{he2020pvn3d} to estimate the 3D keypoint and axis.

\subsubsection{3D Keypoint Estimation}
We sample 3D seed points and estimate the keypoint position by learning the position offset between all the seeds and the target keypoint. Each ROI consists of $N \times N$ superpixels.
The image pixel coordinates and depth values of the seeds are computed through bilinear interpolation similar to RoIAlign~\cite{he2017mask}. Their 3D positions are restored through an inverted camera intrinsic transformation. 

The keypoint loss is defined as L1 loss of estimated 3D offset against ground truth offset filtered by ground truth mask: 
\begin{equation}
\label{eq:voting}
    f_{\text{vote}}(\text{loss}(i)) = \frac{\sum_{i=1}^{N\times{N}} \text{loss}(i) \cdot M_{i}^{*}}{\sum_{i=1}^{N\times N}M_{i}^{*}}
\end{equation}
\begin{equation}
    L_{\text{keypoint}} = f_{\text{vote}}\Bigg(\sum_{k\in\{x,y,z\}}\left \| t_{i}^{k} - t_{i}^{k*} \right\|_{1}\Bigg)
\end{equation}
where Equation~\eqref{eq:voting} describes the voting loss function $f_{\text{vote}}$ which averages the loss value $\text{loss}(i)$ at seed $i$ region with the confidence score $M_{i}^{*}$. $t_{i}^{k}$ denotes the offset of seed $i$ along direction $k$, $t_{i}^{k*}$ denotes ground truth offset. 


\subsubsection{Directed Axis Estimation}

We represent the directed axis using two 3D endpoints. 
We learn two separate 3D keypoints to form the axes using a network structure similar to the 3D keypoint estimation head. 

The loss function is composed of three parts. The first loss $L_{\text{endpoint}}$ is similar to $L_{\text{keypoint}}$, where the loss is the sum of losses for both endpoints. The second loss $L_{\text{axis}}$ encourages the voter points to be near to the axis by calculating the distance from the voter points to the ground truth axis:
%
\begin{equation}
    L_{\text{axis}} = f_{\text{vote}}\left(\left\|(t_{i,m}-t_{i,m}^{*})\times n^{*} \right \|_2\right)
\end{equation}
where $t_{i,m}^{*}$ is the ground truth translation from the seed $i$ to the $m_{th}$ axis endpoint, and $n^{*}$ is the ground truth normalized 3D axis.
The third loss $L_{\text{direction}}$ corrects the direction of the estimated axis. Since the connection of two 3D keypoints determines the axis direction, the directions of connection between corresponding voters of the two keypoints should be close to the ground truth axis:
%
\begin{equation}
L_{\text{direction}}=f_{\text{vote}} \left(1 - \left(t_{i,2}-t_{i,1}\right)\cdot n^{*}\right)
\end{equation}

The final loss for axis estimation is the sum of $L_{\text{endpoint}}$, $L_{\text{axis}}$ and $L_{\text{direction}}$. During inference, two endpoints are clustered with the same method as for 3D keypoint estimation. Then, the final output axis is a directed link between endpoints.

\subsubsection{Part Affinity Field Estimation}
In multi-instance environment, we also need to determine which parts can be associated to the whole objects. In human pose estimation community, to represent association property between joints, Part Affinity Field (PAF) was proposed in~\cite{cao2017realtime}. PAF is a set of 2D unit vectors that indicate the connecting directions between neighbor limbs. Inspired by this, the network is designed to predict a set of 2D unit vectors, which determine the potential associated part direction for each ROI. The network structure is similar to the mask head, estimating 2D vectors $p_{i}$ for seed $i$ that point from the corresponding part keypoint to its associated part keypoint. 
The loss function is defined as:

\begin{equation}
    L_{\text{paf}} = f_{\text{vote}} (\left \|p_{i}-p^{*} \right \|)
\end{equation}
where $p^{*}$ is the ground truth 2D unit vector pointing toward the target part keypoint. During inference, the mean direction of affinity fields within estimated mask will be used as final field direction.

\section{Experiments}

We evaluate our method from the following perspectives: part-level pose estimation accuracy; and
robustness of applying ACF to robot manipulation.

\begin{figure}
    \centering
    \includegraphics[width=\columnwidth]{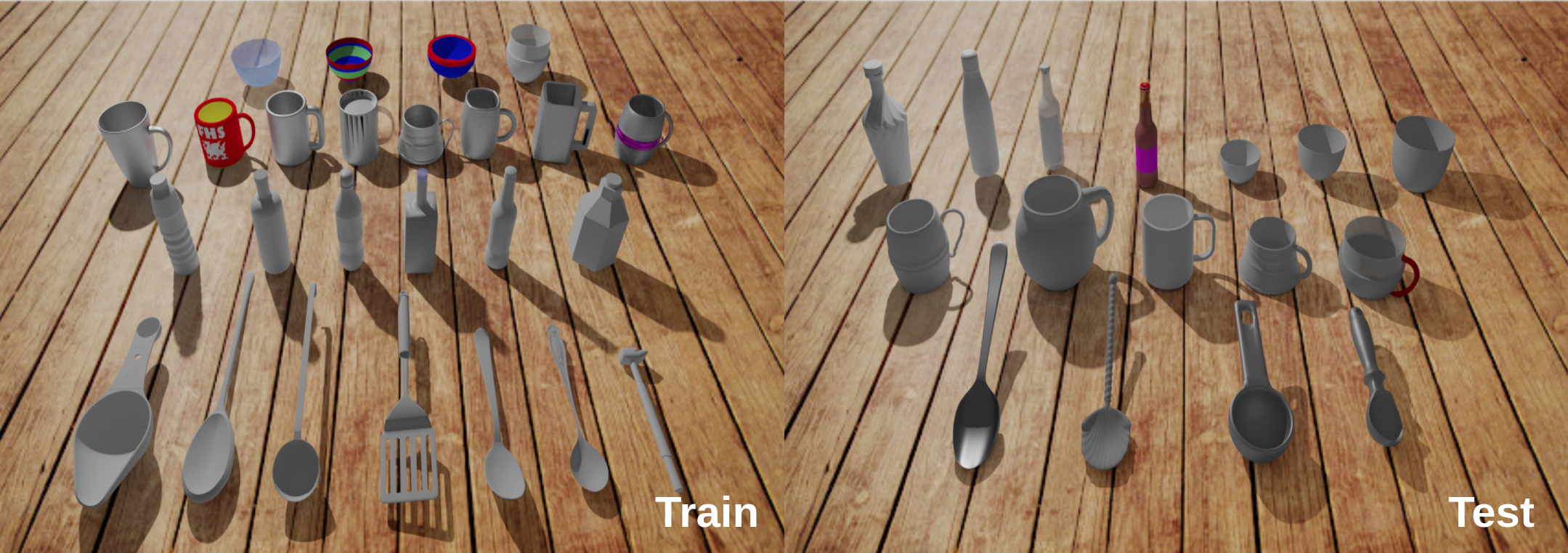}
    \caption{The models in the train set and the validation / test set. For the synthetic dataset and robot simulation experiments, we add random texture to each object, as shown in Figure~\ref{fig:acf_manipulation1} and Figure~\ref{fig:acf_manipulation2}.}
    \label{fig:dataset}
\end{figure}

\textbf{Dataset:} 
We create a synthetic dataset of images which includes RGB, depth, instance segmentation and object part pose in cluttered indoor environments using the NDDS data generation plugin~\cite{to2018ndds} in Unreal Engine 4. We select several hand-scale objects that support robot grasping and manipulation for the drink-serving task. The object models include the bottle, mug and bowl categories from the ShapeNet dataset~\cite{shapenet2015}, and the spoon, spatula, hammer categories from public available object model cites like TurboSquid~\cite{turbosquid}, as shown in Figure~\ref{fig:dataset}. 
We divide objects into four predefined parts: \textit{container}, \textit{handle}, \textit{stir}, \textit{scoop}. The object-part-action relation is detailed in Table.~\ref{tab:object-affordance}. The keypoints for \textit{container} and \textit{scoop} are defined as the center of their geometry shapes. The keypoints for \textit{scoop} and \textit{handle} are the tangent points on the line parallel to the upright orientation. The axes are labeled as follows: \textit{container}, from bottom to top; \textit{handle}, from external tangent point to container direction; \textit{handle}, from the tail endpoint to scoop endpoint; \textit{scoop}, the upwards normal at the center point of concave surface. Examples are shown in Fig. \ref{fig:acf_explain}.
During the rendering, all object models are randomly translated and rotated.
Scene backgrounds and object textures are randomized. In total, we render 20k images, with 1k images each for validation and test.

\textbf{Implementation:} We use Pytorch to implement the deep network modules. We use ResNet50~\cite{he2016deep} with feature pyramid network as a backbone with a pretrained model on the COCO dataset. The initial weight for the input depth channel is set as the average of the 3 RGB channels at the first layer of the backbone network. The axis endpoint head and the 3D keypoint head consist of shared multi-layered perceptrons.
During inference, from each ROI, we select seed samples whose projected coordinates fall into masks, and get 3D position voters by adding estimated offsets. Then, the Mean Shift algorithm~\cite{comaniciu2002mean} is applied to cluster voters, and the center of clusters is regarded as the final estimate. The implementation of the mean shift algorithm for deep Hough voting has the same structure as in He et al.~\cite{he2020pvn3d}. We trained our network on a Nvidia 2080 Super GPU with a batch size of 2 for 60 epochs, using the Adam optimizer with 0.001 learning rate and 0.001 weight decay.


\begin{table}[t]
\centering
\caption{Object-Part-Action Relations. The top table indicates the parts contained in an object with an `x'. The bottom table indicates the actions which correspond to a part. The `container' and `handle' parts are separated in the way shown in Figure~\ref{fig:acf_explain}. `stir' refers to the straight handle part of spoons, spatulas and hammers, and `scoop' refers to the head part that supports scoop of spoons and spatulas.}\label{tab:object-affordance}
\footnotesize
\begin{tabular}{r c c c c}
\toprule
 & \multicolumn{4}{c}{\textbf{Part}} \\  \cmidrule{2-5}
\textbf{Object} & container & handle & stir & scoop \\ 
\midrule
bottle & x      &     &  &  \\ 
mug   & x      & x   &  &      \\ 
bowl  & x      &     &  &       \\ 
spoon   &       &    & x &   x   \\ 
spatula &       &    & x &   x   \\ 
hammer  &        &    & x &       \\ \midrule
\textbf{Action} & container & handle & stir & scoop \\ \midrule
grasp   & x      & x   & x &      \\ 
stir  &       &     & x &       \\ 
scoop  &       &     &  x &   x    \\ 
contain  &   x     &    &   &      \\ 
pour  &   x     &    &   &       \\ 
\bottomrule
\end{tabular}
\end{table}

\subsection{Ablation Studies on Axis Estimation}
\label{par:ablation}
Apart from using two 3D keypoints as representation (we call the method \textbf{ACF-Endpoints}), we tested another two ways to represent a 3D axis. \textbf{ACF-Vector} represents the ACF as a 3D unit vector. 
During training, the loss $L_{\text{vector}}$ is defined to be the difference between estimated vector $n$ and ground truth direction vector $n^*$. During inference the estimated vector is directly regarded as the axis. 
\textbf{ACF-ScatterLine} uses a 3D point set along the axis, with a binary label indicating the closer endpoint for each point in the set. During training, the loss is defined as the sum of $L_{\text{axis}}$, $L_{\text{inner}}$ and $L_{\text{label}}$. $L_{\text{inner}}$ pushes points to make offsets $t$ perpendicular to the axis direction $n^*$ so that points spread out along the axis direction which benefits the linear regression during inference. $L_{\text{label}}$ corrects the two endpoints order by a binary cross entropy with logits loss (BCELoss in Equation (8)) between estimated closer endpoint index $l$ and ground truth closer endpoint index $l^*$. During inference, the axis is solved through linear regression with RANSAC from estimated point set and the axis direction is determined by point labels.

\begin{equation}
L_{\text{vector}}=f_{\text{vote}} (\left \|n_i - n^{*} \right\|)
\end{equation}
\begin{equation}
L_{\text{inner}}=f_{\text{vote}} (t_{i}\cdot n^{*})
\end{equation}
\begin{equation}
L_{\text{label}}=f_{\text{vote}} (\text{BCELoss}(l_i,l_i^*))
\end{equation}



\begin{figure*}
    \centering
    \includegraphics[width=0.95\textwidth]{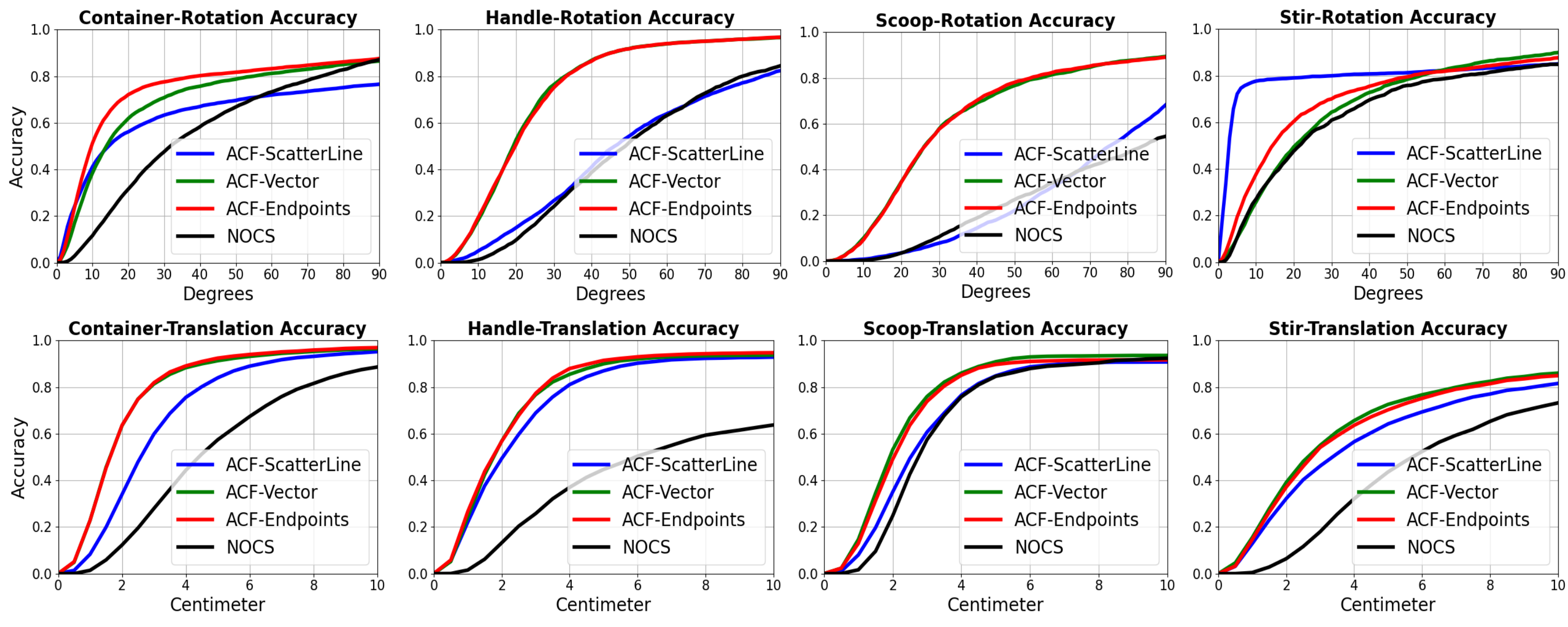}
    \caption{Comparison between each method for different part estimation with respect to mAP curves. The top row shows rotation accuracy and the bottom row shows translation accuracy.}
    \label{fig:result}
\end{figure*}

We compare the three variants of learning ACF axis. The metric of calculating angular distance between estimated and ground truth axis $n_1, n_2$ is defined as $180 / \pi \cdot\arccos{(n_1\cdot n_2)}$. From Table~\ref{tab:comparenocs} and the first row in Figure~\ref{fig:result}, we find ACF-Endpoints achieves the highest mean average precision (mAP) on \textit{container}, \textit{handle} parts and slightly worse as ACF-Vector on \textit{scoop} parts. The reason might be that the \textit{scoop} parts do not span a long distance along the direction of defined ACF axis, so the two endpoints cannot be learned to be as distinct as other parts. ACF-ScatterLine performs much better in \textit{stir} and \textit{container} parts compared to \textit{scoop} and \textit{handle}. We believe that is because the line-shape scatter suits better to straight line or cylinder shapes with a clear center axis or a distinct primary direction. Overall, ACF-Endpoints performs robustly among parts with different shapes and achieves 40.9 mAP within 15$^\circ$, 5cm threshold.

\begin{table}[t]
\centering
\caption{Comparison of proposed ACF method with different axis representation and NOCS on object part keypoint position and axis estimation. The numbers inside the table are mean Average Precision (mAP) within different error tolerances.}
\label{tab:comparenocs}
\footnotesize
\setlength{\tabcolsep}{0.8mm}{
\begin{tabular}{llcccc}
\toprule
\textbf{Part}   & \textbf{Method} & 10$^{\circ}\vert$2cm & 15$^{\circ}\vert$2cm & 10$^{\circ}\vert$5cm & 15$^{\circ}\vert$5cm \\ \midrule
\multirow{4}{*}{container} & NOCS & 4.9 & 7.2 & 8.9 & 15.9 \\ 
    &  ACF-Endpoints &  \textbf{37.7}  & \textbf{46.3} &  \textbf{49.7}  & \textbf{63.1} \\
    & ACF-Vector &  28.4  & 37.8 &  37.4  & 51.0 \\
    & ACF-ScatterLine  &  17.0  & 20.8 & 36.0  & 44.7 \\
\midrule
\multirow{4}{*}{handle} & NOCS & 0.4 & 1.7 & 0.7 & 2.6 \\ 
    &  ACF-Endpoints   &  \textbf{14.4}  &  \textbf{24.9} & \textbf{19.2} &  \textbf{35.0}  \\
    & ACF-Vector &  12.1  & 22.4 &  17.7  & 33.6 \\
    & ACF-ScatterLine  &  3.0  & 6.1 &  1.4  & 4.7 \\
\midrule
\multirow{4}{*}{stir} & NOCS  & 3.1 & 4.1 & 17.9 & 24.1\\ 
    &  ACF-Endpoints   &  20.4 & 26.4 &  33.3  & 45.0 \\
    & ACF-Vector  &  15.9  & 23.2 &  23.9  & 35.8 \\
    & ACF-ScatterLine  &  \textbf{30.2}  & \textbf{30.5} &  \textbf{58.3}  & \textbf{58.8} \\
\midrule
\multirow{4}{*}{scoop} & NOCS & 0.0 & 0.1 & 0.1 & 0.3\\ 
    &  ACF-Endpoints   & 7.1 &  14.9  &  9.2  & 20.4 \\
    & ACF-Vector  &  \textbf{7.5}  & \textbf{15.9} & \textbf{10.0}  & \textbf{20.9} \\
    & ACF-ScatterLine  &  0.3  & 0.7 &  0.7  & 1.8 \\
\midrule
\multirow{4}{*}{mean} & NOCS & 2.1 & 3.2 & 6.9 & 10.8\\ 
    &  ACF-Endpoints  &  \textbf{19.9}  & \textbf{28.1} &  \textbf{27.9}  & \textbf{40.9} \\
    & ACF-Vector  &  16.0  & 24.8 &  22.2  & 35.3 \\
    & ACF-ScatterLine  &  12.6  & 14.5 &  24.9  & 28.6 \\
\bottomrule
\end{tabular}
}
\end{table}

\begin{figure*}
    \centering
    \includegraphics[width=0.95\textwidth]{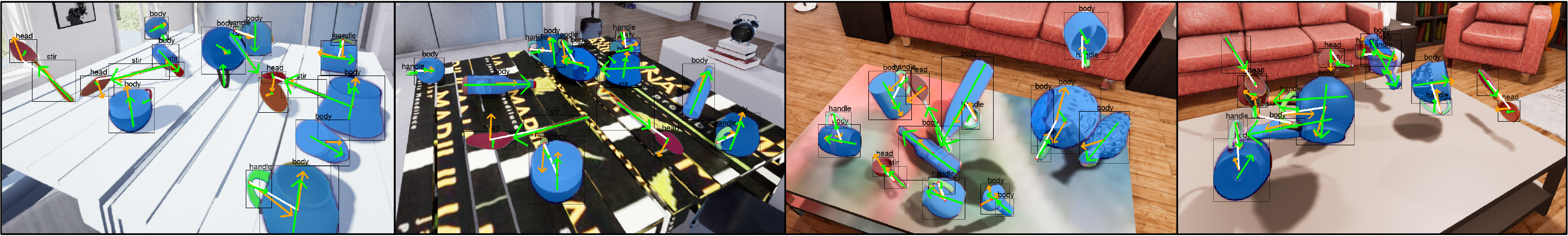}
    \caption{Qualitative results on our dataset with the ACF-Endpoints method. The red points show the keypoint of each part, orange arrows show the estimated axes and green arrows show the ground truth axes. Then, the white line shows the connection of two parts.}
    \label{fig:preception}
\end{figure*}

\subsection{Comparison with NOCS}

We compare our method with a state-of-the-art category-level 6D object pose estimation method \textbf{NOCS}~\cite{wang2019normalized}. 
We trained NOCS network to do part category-level pose estimation and compared the two methods with respect to part keypoint position and axis direction. The results are shown in Table ~\ref{tab:comparenocs} and Figure~\ref{fig:result}. We can figure out that the NOCS does not perform well in our tasks. We think there are two main reasons: (1) The intra-class variance. In the NOCS original paper, they considered object-level classes. For instance, they separate objects into mug, bottle, and bowl classes based on their geometry. However, we regard all of these containers as the container and handle classes based on their functionality. The container parts of mug, bottle and bowl could vary more than that within one object class.  (2) The amount of data versus information needed to learn. Compared to the training dataset used in~\cite{wang2019normalized}, our dataset is about 20X smaller and our training iteration is 4X fewer. 

\subsection{Object-Level Detection Accuracy}

Besides part level keypoint and axis estimation for manipulation, we also aim to evaluate object level detection. The intuition is that compared to whole object detection, part-based representation might be more robust in highly cluttered scenes where objects are partly occluded. We choose mugs as test objects and compare our method with a Mask R-CNN baseline. 
Mask R-CNN achieves 93.2 and 70.6 mAP on uncluttered and cluttered scenes respectively, while ACF-Endpoints achieves 93.3 and 90.2 mAP on uncluttered and cluttered scenes.
Both methods perform similarly in uncluttered scenes, while ACF-Endpoints performs better in heavily clutters. The result supports reliability for robotic manipulation in cluttered environments based on the proposed part-level perception method.

\subsection{Application to Robotic Manipulation Task}

\begin{figure}[htbp]
    \centering
    \includegraphics[width=0.9\textwidth]{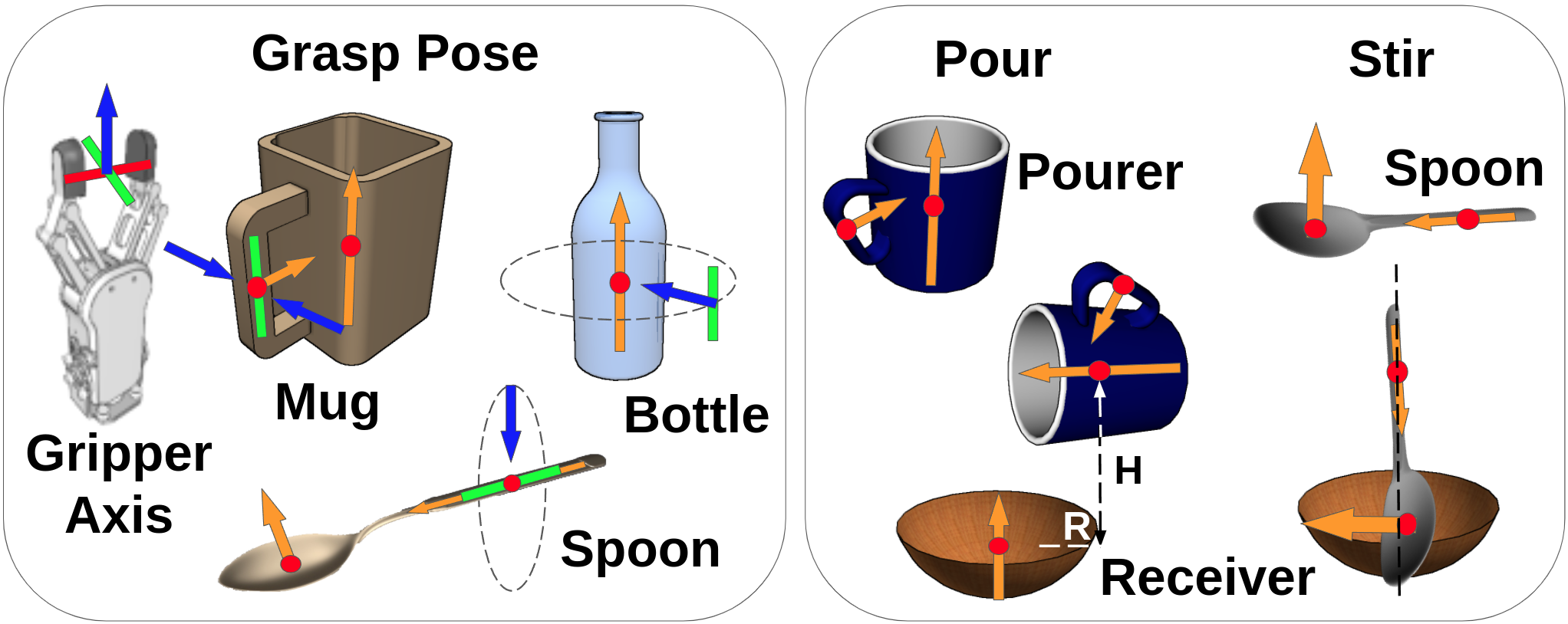}
    \caption{Grasp poses and pour, stir manipulation waypoints composed from detected ACFs.}
    \label{fig:acf_action}
\end{figure}

To test the perceived ACF applicability, we created a scenario for robotic manipulation to complete a drink-serving task in CoppeliaSim simulation platform~\cite{rohmer2013coppeliasim} as well as using a Fetch robot in real world. During the task execution, we assume individual step sequence (grasp, pour, stir) is given as shown in Figure~\ref{fig:acf_action} so that after successful grasps, the pour, stir action can be finished without need of re-grasping. 

For mugs, the grasp pose is located at handle keypoint, with blue axis orthogonal to both the handle and container axes, and green axis parallel to the container axis; For bottles, the grasp pose is located at container keypoint, with green axis parallel to container axis, and blue axis towards the container keypoint; For spoons, the grasp pose is located at handle keypoint, with green axis parallel to its stir axis, and blue axis towards the handle keypoint. We don't have grasp poses for bowls yet.
For pouring, the trajectory of pourer is first constrained to avoid liquid spilling by holding its container part axis to be upright. The pouring action rotates the axis in vertical plane, with keypoint located at a fixed height $H$=15cm, and a rotation radius $R$=5cm relative to the receiver's keypoint. 
For stirring, the spoon is supposed to move downwards with its stir axis aligned with container axis, till its scoop keypoint reaches the container keypoint. Then the spoon is to move along its scoop axis to stir.

\begin{figure*}[htbp]
    \centering
    \includegraphics[width=\textwidth]{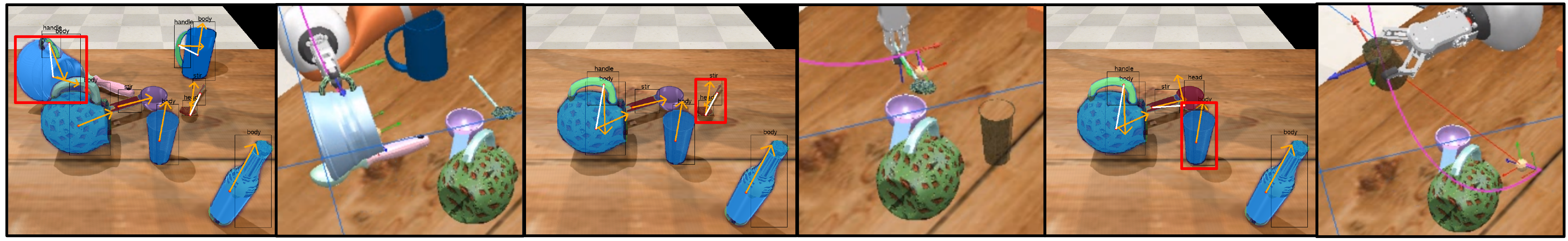}
    \caption{An example of ACF estimation based grasping pipeline. From left to right, 3 pairs of estimation result images (detected body, handle, stir, head are with blue, green, red and purple masks) and grasping snapshots are shown for 3 objects (highlighted in red bounding boxes) in object clutter. The grasp poses are decided as shown in Figure~\ref{fig:acf_action}.}
    \label{fig:acf_manipulation1}
\end{figure*}
\begin{figure*}[htbp]
    \centering
    \includegraphics[width=\textwidth]{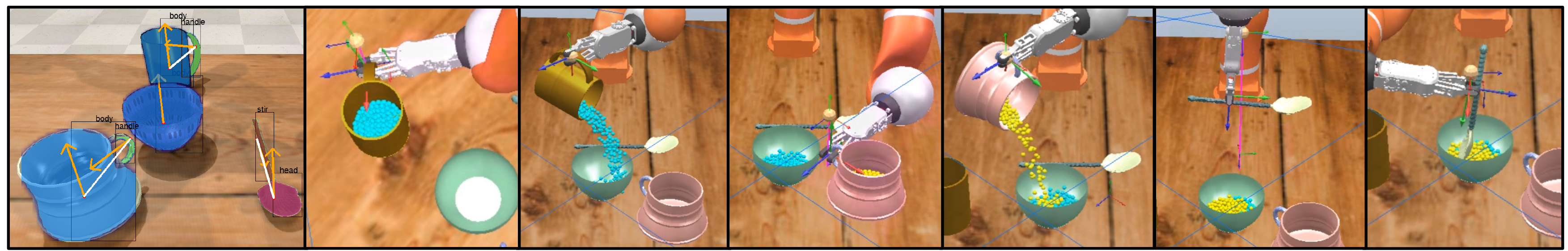}
    \caption{An example of ACF estimation based drink serving pipeline. From left to right, we show the ACF estimation result (left most) and snapshots of grasping two containers and pouring liquid (in cyan and yellow), and grasping a spoon to stir the liquid inside the container.}
    \label{fig:acf_manipulation2}
\end{figure*}
As shown in Figure~\ref{fig:acf_manipulation1},~\ref{fig:acf_manipulation2}, random object clutters including containers and spoons are generated on tabletop, and a KUKA arm equipped with a parallel gripper is performing actions step by step to finish the task. Two RGB-D cameras are set on the two sides of the table and provide input to the ACF perception system, which estimates ACF keypoints and axes. The manipulator IK and motion planning (SBL~\cite{sanchez2003single}) algorithms are provided by CoppeliaSim and OMPL library. 

We evaluate the accuracy of ACF estimation with respect to the task success rate of three individual actions: grasping, pouring liquid (simulated as a set of particles) from one container into another, and stirring liquid in a container. We also evaluate the overall drink serving task, which includes pouring two kinds of liquid into the same container and stir the liquid afterwards.
The grasping is considered successful when the gripper is able to grasp the object and lift it up without dropping. For evaluations on pouring actions, we calculate the ratio of the change in liquid volume of pourer and receiver, and consider it successful if the ratio is greater than 0.7. For stirring, it is considered successful when the position error is less than 2cm. The overall drink serving task is considered successful if all single actions succeed. Number of trials and success rates for all tasks are listed in Table~\ref{tab:experiment}. We find mugs and spoons are more challenging to grasp than bottles, because spoons have larger position errors in ACF estimation and mug handles have very small areas for valid grasps as shown in Figure~\ref{fig:acf_action}. We find similar success rate for pour and stir actions, while pour often fails during the pouring action and stir often fails during motion. The main failure of the pour task is that much of water is poured outside of container because the pourer doesn't reach close to above the center of receiver. On the other hand, the stir task requires the robot to insert the spoon into the container from above, which will lead to collision with container inner surface. For drink serving, many trials fail on one of two pour actions which leads to a lower success rate. We expect taking masked object part point cloud into consideration could help design more adaptive trajectories.

\begin{table}
\caption{Task evaluation on grasp, pour, stir and drink serving which includes two pour and a stir action in sequence.}
\label{tab:experiment}
\footnotesize
\renewcommand{\arraystretch}{1.1}
\begin{tabular}{lcc}
\toprule
\textbf{Task} & \textbf{\# Trials} & \textbf{Success Rate}\\
\midrule
Grasp Mug & 340 & 62.65\%\\
Grasp Bottle & 162 & 70.99\%\\
Grasp Spoon & 152 & 63.81\%\\
Pour & 130 & 78.46\%\\
Stir & 133 & 71.43\%\\
Drink serve & 143 & 37.06\%\\
\bottomrule
\end{tabular}
\end{table}

\begin{figure*}[htbp]
    \centering
    \includegraphics[width=0.9\textwidth]{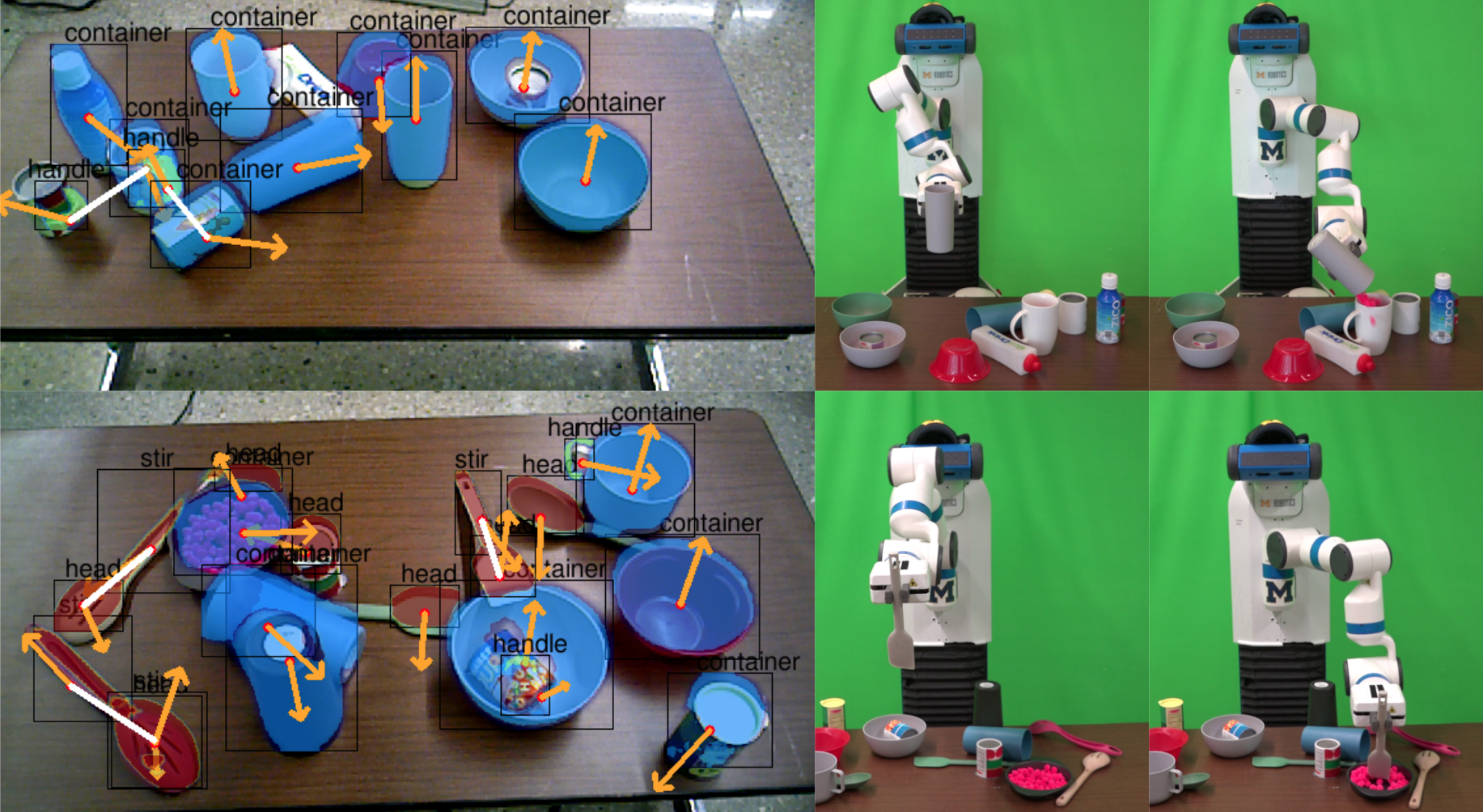}
    \caption{The mask, ACF keypoint and axis estimations are overlaid on the left images. On top-right, Fetch robot is grasping cups, pouring to another container based on ACF estimation in top-left image. On bottom-right, Fetch robot is picking up a spoon to stir inside a bowl based on bottom-left image. }
    \label{fig:fetch_exp}
\end{figure*}

We further evaluated the efficacy of ACF-based manipulation on a real Fetch robot. We realized the network purely trained from synthetic data could not generalize well to real objects, so we fine-tuned the network with around 10K images collected and labelled through ProgressLabeller~\cite{progress2022chen}. We tested pour and stir tasks and got both 60\% success rate out of 10 trials over novel instances. In failed trials, the errors were mostly due to the MoveIt! motion planner.  Figure~\ref{fig:fetch_exp} shows snapshot of one grasp-pour and one grasp-stir task. The entire tasks recording will be attached in the supplementary video.

\section{Conclusion}
We present a novel representation of object affordances called Affordance Coordinate Frame (ACF), and we propose a deep learning based perception method for estimating ACF given a RGB-D image. We demonstrate that our proposed perception method outperforms state-of-the-art method (NOCS~\cite{wang2019normalized}) in estimating ACF of novel objects. Our method also outperforms Mask R-CNN in detecting novel objects especially under cluttered environment. We further demonstrate the applicability of ACF for a drink serving task. In future, we will extend the work to incorporate a larger variety of object categories and compare with other analytical geometry-based manipulation systems.

\balance


\bibliographystyle{IEEEtran}
\bibliography{ref}

\end{document}